\documentclass[letterpaper, 10pt, conference]{ieeeconf}

\usepackage{times}
\usepackage[normalem]{ulem}	                        
\usepackage[table,usenames,dvipsnames]{xcolor}      
\usepackage{extarrows}                              
\let\labelindent\relax
\usepackage{enumitem}                               

\usepackage{amsmath,amsthm,amssymb,amsfonts,dsfont} 
\usepackage{algorithm,algorithmicx,listings}        
\usepackage[noend]{algpseudocode}			        
\usepackage{textcomp}
\usepackage{graphicx}
\usepackage{tabularx}
\usepackage{subcaption}
\usepackage{multirow,multicol, array}
\usepackage[font={small}]{caption}   
\usepackage{tikz}
\usetikzlibrary{matrix,chains,positioning,decorations.pathreplacing,arrows}
\usepackage{balance}

\usepackage[breaklinks=true, colorlinks, bookmarks=true, citecolor=Black, urlcolor=Violet,linkcolor=Black]{hyperref}

\newcommand{\prl}[1]{\mathopen{}\left(#1\right)\mathclose{}}

\newtheorem*{remark*}{Remark}
\newtheorem*{problem*}{Problem}

\makeatletter
\def\maketag@@@#1{\hbox{\m@th\normalfont\normalsize#1}}
\makeatother

\IEEEoverridecommandlockouts
\begin{document}
\title{Neural Network Memory Architectures for Autonomous Robot Navigation}
\author{Steven W. Chen, Nikolay Atanasov, Arbaaz Khan, Konstantinos Karydis, Daniel D. Lee, and Vijay Kumar
\thanks{This work is supported in part by ARL \# W911NF-08-2-0004, DARPA \# HR001151626/HR0011516850, ARO \# W911NF-13-1-0350, and ONR \# N00014-07-1-0829. The authors are with the GRASP Laboratory, University of Pennsylvania. Email: \tt\small\{chenste, atanasov, arbaazk, kkarydis, ddlee, kumar\}@seas.upenn.edu}
}

\maketitle

\begin{abstract}
This paper highlights the significance of including memory structures in neural networks when the latter are used to learn perception-action loops for autonomous robot navigation. Traditional navigation approaches rely on global maps of the environment to overcome cul-de-sacs and plan feasible motions. Yet, maintaining an accurate global map may be challenging in real-world settings. A possible way to mitigate this limitation is to use learning techniques that forgo hand-engineered map representations and infer appropriate control responses directly from sensed information. An important but unexplored aspect of such approaches is the effect of memory on their performance. This work is a first thorough study of memory structures for deep-neural-network-based robot navigation, and offers novel tools to train such networks from supervision and quantify their ability to generalize to unseen scenarios. We analyze the separation and generalization abilities of feedforward, long short-term memory, and differentiable neural computer networks. We introduce a new method to evaluate the generalization ability by estimating the VC-dimension of networks with a final linear readout layer. We validate that the VC estimates are good predictors of actual test performance. The reported method can be applied to deep learning problems beyond robotics.
\end{abstract}

\section{Introduction}
\label{sec:intro}

Autonomous robot navigation in real-world settings involves planning and control with limited information in dynamic and partially-known environments. Traditional approaches close perception-action feedback loops by maintaining a global map representation of the environment and employing feedback motion planning algorithms (e.g., optimization-based~\cite{rimon1992exact,lqr-trees}, search-based~\cite{hart1968formal,likhachev2005anytime}, or sampling-based~\cite{Lavalle98rapidly-exploringrandom,rrt-star,Kobilarov-RSS-11,arslan2013use,prm}). While a global map allows navigation in environments with cul-de-sacs and complex obstacle configurations, maintaining one can be challenging due to localization drift, noisy features, environment changes and limited on-board computation~\cite{liu_high_2016,Thrun02a}.

An alternative approach to global mapping and replanning is to determine a closed-loop policy that maps the history of sensor positions and observations to the next action at each time step. The benefit of computing an action directly from the observation history is that there is no need to maintain a map, and requires just one function evaluation as opposed to a computationally-expensive search or optimization. Unfortunately, the state or action spaces in robotics applications can be very large, to the extent that keeping the measurement history and representing such policy functions efficiently (e.g., as a lookup table or a linear function) is infeasible \cite{Tesauro:1995:TDL:203330.203343}. Due to their representational power, neural-network-based learning techniques have become increasingly of interest to the robotics community as a method to encode such perception-action policies or value functions~\cite{richterlearning,visuomotor}. 
Furthermore, traditional methods for navigating in unknown environments can be used to supervise the training process by providing action labels to the neural network. Recent work has shown that neural networks can be used to navigate a wheeled robot in cluttered environments \cite{pfeiffer2016perception}.

Despite their success in several fields, deep learning techniques have several limitations when considering robot navigation. Current network architectures are not well-suited for long-term sequential tasks \cite{fusi2005cascade,ganguli2008memory}, which typically appear in robotics (e.g., path planning), because they are inherently reactive. More precisely, while maintaining the complete observation history is not feasible, the spatio-temporal correlation present in observation data necessitates the use of memory structures that summarize observed features (e.g., the map plays this role in traditional navigation approaches). However, the effect of memory on neural network performance and generalization ability is not well-understood. Current methods for evaluating network structures are based on empirical performance on a held-out test set. A limitation of this approach is that it is dependent on the test set choice and is prone to over-fitting during the model-selection phase \cite{cawley2010over}. This limitation is especially important in a robotics context where the unknowns of real-world scenarios may not be easily captured by a test set.

This paper investigates the ability of several neural network memory structures to separate action classes and generalize to unseen environments in the context of robot navigation. Motivated by the reliance of traditional approaches on accurate global maps, our work seeks to reveal new directions and limitations of applying deep learning to the robot navigation problem. We argue that understanding the effect of memory on separability and generalization ability is fundamental for successfully applying neural networks to robotics. Our work makes three main contributions.

%

\begin{itemize}
\item We investigate the relationship between memory, separability and generalization ability of neural-network perception-action policies in cul-de-sac environments. 
\item We estimate the Vapnik-Chervonenkis (VC) dimension~\cite{Vapnik_1995} of the last layer of a neural network (after all upstream layer transformations) as a measure of the network generalization ability that depends only on the training set choice.
\item We develop a new parallel training algorithm for supervised learning of perception-action policies in sequential prediction problems.
\end{itemize}

\section{Related Work}
\label{sec:relatedwork}

Recent works have considered the application of deep learning techniques to autonomous robot navigation. 
Tamar et al. developed Value Iteration Networks (VIN) that outperform regular convolutional networks in planning-based sequential reasoning \cite{vi_networks}. However, that work only explored planning in known maps. 
Zhang et al. used Model Predictive Control (MPC) in guided policy search to train a neural network to navigate a quadrotor using lidar inputs \cite{mpc_guided}. Ross et al. used the Dataset Aggregation (DAgger) supervised learning algorithm to navigate in a dense forest environment using image inputs, and concluded that incorporating memory could improve failure cases \cite{ross2013learning}. Existing works consider only convex obstacles, and as a result they may have limited success in autonomous robot navigation in environments which contain complex obstacles such as cul-de-sacs~\cite{dey2016vision}. 


These prior works and observations suggest that some form of memory is necessary. The most commonly used method to incorporate memory in robotics learning problems are through Recurrent Neural Networks (RNN) such as the Long-Short Term Memory (LSTM) \cite{Hochreiter:1997:LSM:1246443.1246450}. Heess et al. demonstrate that recurrent neural networks are able to learn in partially-observable problems \cite{heess2015memory}. Mnih et al. incorporate the LSTM layers into their Asynchronous Advantage Actor-Critic (A3C) algorithm and demonstrate improved performance over a feedforward network \cite{a3c}. Zhang et al. argue that memory is necessary in partially-observed tasks, and augment the state space to include memory states \cite{DBLP:journals/corr/ZhangLMFA15}. 

Graves et al. observe that while RNN's can in principle be used to simulate arbitrary procedures, learning the optimal network is not easy in practice \cite{DBLP:journals/corr/GravesWD14}. This observation inspired a new family of neural network architectures called Memory Augmented Neural Networks (MANN) which utilize an explicit memory matrix \cite{DBLP:journals/corr/RaeHHDSWGL16}. Among them, the Differentiable Neural Computer (DNC) has demonstrated improved performance over standard RNNs in tasks such as copying sequences and planning shortest paths \cite{graves2016hybrid}. However, MANN structures have been mostly applied to natural language processing and speech recognition; their application to robotics has received less attention.

\section{Problem Formulation}
\label{sec:problem}

Consider a bounded connected set $\mathcal{X}$ representing the workspace of a robot. Let $\mathcal{X}^{obs}$ and $\mathcal{X}^{goal}$, called the obstacle region and the goal region, respectively, be subsets of $\mathcal{X}$. Denote the obstacle-free portion of the workspace as $\mathcal{X}^{free} := \mathcal{X}\backslash\mathcal{X}^{obs}$. The dynamics of the robot are specified by the Probability Density Function (PDF) $p_f(\cdot \mid x_t, u_t)$ of the robot state $x_{t+1} \in \mathcal{X}$ at time $t+1$ given the previous state $x_t \in \mathcal{X}$ and control input $u_t\in \mathcal{U}$. We assume that the control input space $\mathcal{U}$ is a finite discrete set.\footnote{For instance, the control space $\mathcal{U}$ for a differential-drive robot in $SE(2)$ can be a set of motion primitives, parameterized by linear velocity, angular velocity and duration. For a quadrotor, $\mathcal{U}$ may be a set of short-range dynamically feasible motions.} 
The robot perceives its environment through observations $z_t \in \mathcal{Z}$ generated from a depth sensor (e.g., lidar, depth camera), whose model is specified by the PDF $p_h(\cdot \mid \mathcal{X}^{obs},x_t)$. The information available to the robot at time $t$ to compute the control input $u_t$ is $i_t := (x_{0:t},z_{0:t},u_{0:t-1},\mathcal{X}^{goal}) \in \mathcal{I}$, consisting of current and previous observations $z_{0:t}$, current and previous states $x_{0:t}$ and previous control inputs $u_{0:t-1}$
\begin{figure}[t!]
\vspace*{0.1in}
\centering
\begin{subfigure}{0.48\textwidth}
\centering
\includegraphics[width=116px]{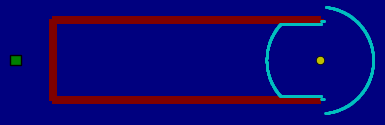}
\includegraphics[width=116px]{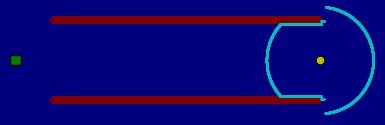}
\end{subfigure}
\vspace{4pt}
\caption{\textbf{\textit{Left}}: Cul-de-sac. \textbf{\textit{Right}}: Parallel Walls. The type of obstacle is unknown until the robot reaches the end of the obstacle. } 
\label{fig:U-map}
\vspace{-8pt}
\end{figure}

\begin{problem*}
\vspace{-3pt}
\label{prob1}
Given an initial state $x_0 \in \mathcal{X}^{free}$ and a goal region $\mathcal{X}^{goal} \subset \mathcal{X}^{free}$, find a function $\mu: \mathcal{I} \rightarrow \mathcal{U}$, if one exists, such that applying the control $u_t := \mu(i_t)$ results in a sequence of states that satisfies $\{x_0,x_1,\ldots,x_T\} \subset \mathcal{X}^{free}$ and $x_T \in \mathcal{X}^{goal}$.
\vspace{-3pt}
\end{problem*}

In this problem setting, the obstacle region $\mathcal{X}^{obs}$ is a partially observable state. Instead of trying to estimate it using a mapping approach, our goal is to learn a policy $\hat{\mu}$ that maps the sequence of sensor observations $z_0,z_1,\ldots$ directly to control inputs for the robot. The partial observability requires an explicit consideration of \textit{memory} in order to learn $\hat{\mu}$ successfully. A partially observable problem can be represented via a Markov Decision Process (MDP) over the information space. More precisely, we consider a finite-horizon discounted MDP defined by $(\mathcal{I}, \mathcal{U}, \mathcal{T}, \mathcal{R}, \gamma)$, where $\gamma \in (0, 1]$ is a discount factor, $\mathcal{I}$ is the state space, $\mathcal{U}$ is the action space, $\mathcal{T}:\mathcal{I} \times \mathcal{U} \times \mathcal{I} \rightarrow [0,1]$ is the transition function, and $\mathcal{R}:\mathcal{I}\times \mathcal{U} \times \mathcal{I} \rightarrow \mathbb{R}$ is the reward function. The latter two are defined as follows:
\begin{align*}
\mathcal{T}(i_{t}^{(1)}, u, i_{t+1}^{(2)}) &:= p_h(z_{t+1}^{(2)} \mid \mathcal{X}^{obs},x_{t+1}^{(2)}) p_f(x_{t+1}^{(2)} \mid x_{t}^{(1)},u)\\
\mathcal{R}(i_{t}^{(1)}, u, i_{t+1}^{(2)}) &:= \begin{cases}
1, & \text{if}~x_{t+1}^{(2)} \in \mathcal{X}^{goal}\\
-1, & \text{if}~x_{t+1}^{(2)}\in \mathcal{X}^{obs}\\
0, & \text{otherwise}
\end{cases}
\end{align*}


In the rest of the paper we consider a 2-D grid world, an instance of the feasible planning problem in which $\mathcal{X} \subset \mathbb{R}^2$ and $\mathcal{U}:=\{down,right,up,left\}$. To investigate the need for memory, we consider U-shape cul-de-sac maps, illustrated in Fig.~\ref{fig:U-map}. The traditional approach to the feasible planning problem in this setting is simultaneous mapping and planning. In contrast, we consider learning a feasible policy by using the outputs of an $A^*$ path planner~\cite{ara_star} for supervision. Let $q_t:=\prl{x_t,z_t,u_{t-1},\mathcal{X}^{goal}}$ be the information available at time $t$ and decompose the information state as $i_t = q_{0:t}$. Our idea is to rely on a neural network to estimate a feasible control policy $\hat{\mu}$ that maps the current information $q_t$ to a control input $u_t$, by computing a $|\mathcal{U}|$-dimensional probability distribution over the possible controls and returning the maximum likelihood control. Such a network needs a hidden memory state $h_t$ in order to remember past information $q_{0:t-1}$ and represents the policy $\hat{\mu}(q_t,h_t;\theta)$ via parameters $\theta$. Our goal is to optimize the network parameters $\theta$ in order to match the output of simultaneous occupancy grid mapping~\cite[Ch.9]{ProbabilisticRoboticsBook} and $A^*$ planning.

\section{Memory Architecture}
\label{sec:memory}

\begin{figure}[t!]
\vspace{0.05in}
\centering
\begin{subfigure}{0.5\textwidth}
\centering
\includegraphics[width=180px]{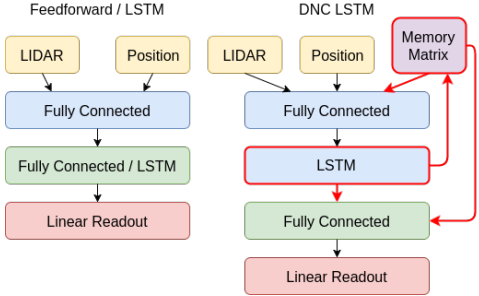}
\end{subfigure}
\vspace{8pt}
\caption{\textbf{\textit{Left}}: We evaluate 4 architectures: FF, LSTM, DNC LSTM, and \textit{regularized} DNC LSTM. The regularized parameters that are regularized are indicated in red. The colors indicate: (yellow) input; (blue) hidden layers, (green) last upstream layer; (red) linear readout layer; and (purple) the memory matrix.} 
\label{fig:network_architectures}
\vspace{-8pt}
\end{figure}

We describe three neural network architectures that use different structures to represent the memory state $h_t$: feedforward (FF), long short-term memory (LSTM) and differentiable neural computer (DNC)---see Fig. \ref{fig:network_architectures}.

\textbf{\textit{Feedforward}}: Deep FF networks, such as the multilayer perceptron~\cite[Ch. 6]{Goodfellow-et-al-2016} and ConvNets, have been very successful in various vision and robotic tasks \cite{krizhevsky2012imagenet}\cite{levine2013guided}. Our first architecture uses only FF layers to model the mapping from inputs $q_t$ to actions $u_t$. This structure does not include a memory state $h_{t}$, and the policy can be rewritten as $\hat{\mu}(q_{t};\theta)$. 

We hypothesize that the lack of a memory state will be a problem in the cul-de-sac environment because there exist two different states $i_{t}^{(1)}\!\!=\!q_{0:t}^{(1)}$ and $i_{t}^{(2)}\!\!=\!q_{0:t}^{(2)}$ (when the robot is entering and exiting the cul-de-sac) such that $\mu_{t}(i_t^{(1)})\!\! \neq \!\!\mu_{t}(i_t^{(2)})$ but $q_{t}^{(1)}\!\! = \!q_{t}^{(2)}$. In other words, the expert (i.e. the $A^*$ planner) maps the same input $q_t$ to two different actions, depending on the history $q_{0:t-1}$, but the FF network will not be able to distinguish this based only on $q_t$.



\textbf{\textit{Long Short-Term Memory}}: The second architecture we consider remedies the memory problem by introducing a long-short term memory (LSTM) layer~\cite[Ch. 10]{Goodfellow-et-al-2016}, which contains an internal state $h_t$. The LSTM has a memory cell that removes and adds information to the hidden state $h_t$ based on the previous state $h_{t-1}$ and the current input $q_t$ using the input, forget and update gates. 

The hidden recurrent state can thus be seen as a form of implicit memory since it can read from the inputs (the tape) to modify its internal state, but cannot write to the tape to affect future decisions. We hypothesize that the addition of this memory feature is necessary since, for example, the previous entering and exiting scenario where $\mu_{t}(i_t^{(1)})\!\! \neq \!\!\mu_{t}(i_t^{(2)})$ and $q_{t}^{(1)}\!\! = \!q_{t}^{(2)}$, will not be a problem for the LSTM network because $h_{t}^{(1)}\!\! \neq \!h_{t}^{(2)}$.



\textbf{\textit{Differentiable Neural Computer}}: Our third architecture uses a more explicit representation of memory $h_{t}$ in the form of a memory matrix, which may provide better memory features to separate the action classes. 
The use of external memory in neural network architectures was inspired by Turing Machines~\cite{DBLP:journals/corr/GravesWD14}. Whereas the LSTM can only read from the tape, the DNC is similar to a Turing Machine in that it can both read and write to the tape (now the inputs and the memory matrix) to modify its internal state. An external memory architecture has been shown to improve performance in natural language processing and other fields, but such an architecture has never been considered in robotics. We expect that it would be very useful for navigation where long sequences of actions may have to be backtracked.

The neural network reads from the memory matrix by first using a set of $R$ read heads to get $R$ read vectors $r_{t-1}^1, \dots, r_{t-1}^R$. These read vectors are then concatenated with the normal inputs $(x_{t}, z_{t})$ and fed through the neural network computation graph. The neural network outputs a vector $v_{t}$, and an interface vector $\xi_{t}$. The interface vector $\xi_{t}$ is used to update the memory matrix in a write update and read update step. After the memory matrix is updated, the neural network reads from the memory matrix again to get $R$ read vectors $r_{t}^1, \dots, r_{t}^R$. These read vectors are combined with $v_{t}$ via a fully connected layer to get the final action output. The weights that determine the interface vector $\xi_{t}$ are the parameters to be learned.\footnote{See \cite{graves2016hybrid} for details on the computations and memory matrix updates.} We also consider a regularized version of the DNC architecture, where we incorporate an $L_{2}$ weight penalty to the parameters that correspond to memory (external memory and LSTM). 

\section{Asynchronous DAgger}
\label{sec:asyncdagger}


This section describes how we optimize the parameters $\theta$ of the networks representing the policy $\hat{\mu}(q_t,h_t;\theta)$. In sequential prediction problems, a common concern is the temporal dependence of the outputs $u_t$ on the previous inputs $q_{0:t}$. This correlation poses a problem in stochastic gradient descent methods, as the gradient estimated from the most recent steps may no longer be representative of the true gradient. Further, the difference between the state distributions between the expert and the learner is a common concern in sequential prediction problems. A naive implementation of supervised learning will have poor performance because the states and observations encountered by the expert will be different than those encountered by the policy \cite{ross2011reduction}. 

The DAgger \cite{ross2011reduction} algorithm addresses both of these problems. At each training iteration, the current policy collects a set of new trajectories and aggregates it to a replay data set. The intuition behind this algorithm is that rather than training on a distribution of states that the expert encounters, by sampling with the previous and current policies, the next policy is being trained on a distribution of states that the policy is likely to encounter. In addition, this sampling from an aggregated dataset reduces correlation. 


We develop an \emph{asynchronous} variant of the DAgger algorithm that breaks correlation through asynchronous gradient updates estimated from independent parallel learners. Asynchronous DAgger is inspired by the Asynchronous Advantage Actor Critic (A3C) algorithm \cite{a3c}, but differs since the A3C algorithm is an actor-critic reinforcement learning algorithm, while ours is a supervised sequential prediction algorithm. Similar to how our treatment of correlated training data is the supervised analogue to the A3C reinforcement algorithm, the original DAgger algorithm is analogous to the original Deep Q-Network (DQN) algorithm \cite{dqn} in that both store an experience replay databank. The pseudo-algorithm \ref{alg:algorithm1} describes the developed Asynchronous DAgger algorithm.


\begin{algorithm}[h]
\caption{Asynchronous DAgger (for each learner thread)}\label{alg:algorithm1}
\begin{algorithmic}[1]
\small
\State // Assume global shared parameter vector $\theta$
\State Initialize global shared update counter $J \gets 0$
\State Initialize thread update counter $j \gets 1$
\State Initialize thread episode counter $t \gets 1$
\State Initialize thread dataset $\mathcal{D} \gets \emptyset$
\State Initialize thread network gradients $d\theta \gets 0$
\Repeat
\Repeat
\State  Observe $q_{t}$
\State  Execute action $u_{t}$ sampled from current global 
\State \hskip\algorithmicindent action policy $\hat{\mu}_{J}(q_{t}, h_{t}; \theta)$
\State	Retrieve optimal action $\mu(i_t)$ from expert
\State \hskip\algorithmicindent and convert to standard basis vector $e_{\mu}$
\State	Add $(\hat{\mu}_{J}(q_{t}, h_{t}), e_{\mu})$ to $\mathcal{D}$
\State $j \gets j + 1$
\Until{terminal $i_{t}$ or $j == j_{max}$}
\For{$(\hat{\mu}_{J}(q_{t}, h_{t}), e_{\mu}) \in \mathcal{D}$}
\State Accumulate gradients wrt $\theta$: $d\theta \!\gets\! d\theta + \frac{d\mathcal{H}(\hat{\mu}_{J}, e_{\mu})}{d\theta}$
\State \hskip\algorithmicindent where $\mathcal{H}(\cdot, \cdot)$ is the cross-entropy loss
\EndFor
\If {terminal $i_{t}$ or $t==t_{max}$}
	\State{Reset episode}
    \State{$t \gets 1$}
\EndIf
\State Perform asynchronous update of $\theta$ using $d\theta$
\State Reset thread update counter $j \gets 1$
\State Reset thread dataset $\mathcal{D} \gets \emptyset$
\State Reset thread network gradients $d\theta \gets 0$
\State $J \gets J + 1$
\Until{$J > J_{max}$}
\end{algorithmic}
\end{algorithm} 

Each parallel learner executes actions in the simulation, estimates a gradient calculated from its most recent actions, and applies that gradient asynchronously to the global neural network. Note that in Asynchronous DAgger, the state distribution is determined by the current policy $\hat{\mu}_{J}$ as opposed to the optimal policy $\mu$. However, also notice that rather than accumulating a dataset and sampling randomly from it, each thread has its own dataset $\mathcal{D}$ consisting of its previous $t$ examples, and this dataset $\mathcal{D}$ is reset after applying the aynchronous gradient. As a result, Asynchronous DAgger encounters the state distribution of its current policy, as opposed to a mixture of its current and previous policies. This algorithm extends the exploitation of parallel asynchronous learners to supervised sequential prediction problems.

\section{Generalization Ability}
\label{sec:generalizability}

We present a new technique for determining the efficacy of various neural network architectures. The technique is based on the maximum margin theory of generalization for Support Vector Machines (SVM). Given a network, we estimate the VC-dimension of a similar architecture that combines the original network with an SVM as the final readout layer. This estimate is calculated using \textit{only training data} and can be used as an alternative to held-out test sets. 

The architecture of most deep neural networks can be broken into two parts: a hierarchy of upstream layers followed by a single readout layer \cite{PhysRevE.93.060301}. The neural network can thus be viewed as a perceptron in the feature space learned by the upstream layers. A good neural network architecture contains upstream layers that effectively ``separate" the various classes with a large margin, which the last linear readout layer can then easily classify. One specific form of perceptron, the SVM, can be used to evaluate the generalization ability of the neural network. Previous works have recognized the benefit of applying maximum margin SVMs to neural networks by using them as the final readout layer in tasks such as image recognition \cite{zhong_ghosh_2000,collobert2004gentle,nagi2012convolutional,tang2013}. These works offer small, but consistent improvement over standard softmax layers.

We employ the SVM to evaluate the generalization ability of a trained neural network. This is achieved by estimating the VC-dimension of the neural network with the final readout layer replaced by a linear SVM. 
Consider a binary classification problem and let
\[ \Psi(q_{i}, h_{i}) = (\psi_{1}(q_{i}, h_{i}), \dots, \psi_{D}(q_{i}, h_{i})) \]
be a vector in feature space of dimension $D$, and $\boldsymbol{w} = (w_{1}, \dots, w_{D})$ be the vector of weights determining a hyperplane in this space. We use $i$ instead of $t$ to emphasize that all actions are aggregated into one data set, effectively ignoring the temporal nature of the sequential prediction. The VC-dimension $\eta$ of the maximal margin hyperplanes separating $N$ vectors $\Psi(q_{1}, h_{1}) \dots \Psi(q_{N}, h_{N})$ can be estimated by 
\[ \eta_{est} = R^{2}|\boldsymbol{w}_{0}|^{2}\enspace, \] 
where $R$ is the radius of the smallest sphere that contains $\Psi(q_{i}, h_{i})$ for all $i$ and $|\boldsymbol{w}_{0}|$ is the norm of the optimal weights \cite{Vapnik_1995}. The norm of the weights is related to the optimal margin $\Delta = \frac{1}{|\boldsymbol{w}_{0}|}$. Thus good generalizability (low VC-dimension) occurs when the margin $\Delta$ is large with respect to  $R$.\footnote{We refer the reader to \cite{Vapnik_1995} and \cite{Vapnik1998} for in-depth details and proofs.}

Calculating the margin $\Delta$ is trivial based on the norm of the weights $\boldsymbol{w}$ and comes standard with most SVM packages. The radius $R$ can be found by using a simple quadratic program solved through standard convex optimization packages. 

Given the set of $N$ vectors $(\Psi(q_{1}, h_{1}) \dots \Psi(q_{N}, h_{N}))$ in $D$-dimensional feature space, define the $D\times N$ matrix $C:=(\Psi(q_{1}, h_{1}), \dots \Psi(q_{N}, h_{N}))$, consider the quadratic program 
\begin{equation*}
\begin{aligned}
& \underset{z}{\text{minimize}}
& & p^{T}C^{T}Cp - \sum_{i=1}^{N}\Psi(q_{i}, h_{i})^{T}\Psi(q_{i}, h_{i})p_{i} \\
& \text{subject to}
& & \sum_{i=1}^{N}p_{i} = 1 \\
& & & p_{i} \geq 0 \ \forall i,
\end{aligned}
\end{equation*}
and let $p^{*} = (p_{1}^*, \dots, p_{n}^*)$ be some optimal solution. The vector $\Psi^* = \sum_{i=1}^{n}\Psi(q_{i}, h_{i})p_{i}^*$ is the center of the smallest enclosing sphere and the squared radius $R^{2}$ is the negative value of the objective function at $p^*$ \cite{schonherr2002quadratic}. The upstream layers of a neural network thus learn the function $\Psi$ which transforms the raw vectors $(q_{i}, h_{i})$ from the original feature space into vectors in the new feature space $\Psi(q_{i}, h_{i})$. The linear readout layer learns a hyperplane with weights $\boldsymbol{w}$ in this feature space, however $\boldsymbol{w}$ is not necessarily optimal. 

To estimate the VC-dimension $\eta_{est}$ of the optimal hyperplane $\boldsymbol{w}_{0}$ we calculate the margin of the optimal hyperplane and radius of smallest bounding sphere on the training data set in this new feature space. Thus, a better neural network learns a $\Psi$ that results in a lower $\eta_{est}$, so that the linear readout layer separates the classes with a large margin. 

The action policy network takes the form $\hat{\mu}(q_{i}, h_{i}) := \sigma(A\Psi(q_{i}, h_{i})+b)$ where $A$ is a matrix representing the weights of the linear readout layer, $b$ is the bias, and $\sigma$ is the softmax function to turn it into a probability vector. This technique can be applied to analyze linear readout neural networks in any classification problem such as an image recognition or text classification. However, the analysis in our problem is further complicated since we are performing sequential prediction, and it is unclear what data set $\mathcal{D}$ we need to estimate $\eta_{est}$ from. We cannot use the current policies $\hat{\mu}_{J}$ to generate this data set, because each network will generate a different data set and the $\eta_{est}$ of each network will not be comparable. As a result, we instead generate the data set by executing the optimal policy $\mu$. 

This technique is a new application of maximum margin theory to estimate the generalization ability of neural networks. Compared to traditional empirical test accuracies, the reported method yields a better estimate of generalization ability for two reasons. 
First, it explicitly defines the generalization ability as opposed to other proxy performance measures. Second, it is not dependent on the choice of the test set. This method can be combined with traditional empirical test measures in any deep learning problem to yield better insight into the performance of a network.

\section{Results and Analysis}
\label{sec:resultsanalysis}

\begin{figure*}[thpb]
\centering
\vspace*{0.05in}
  \includegraphics[width=122px]{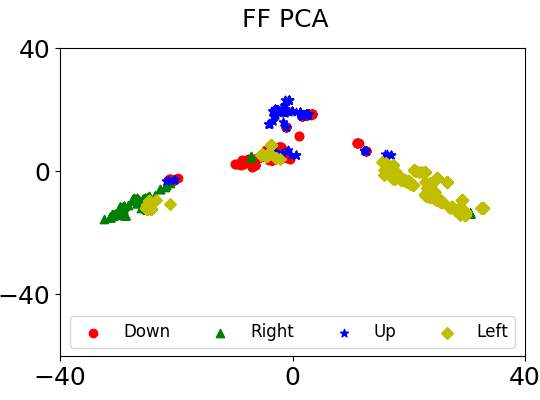}
  \includegraphics[width=122px]{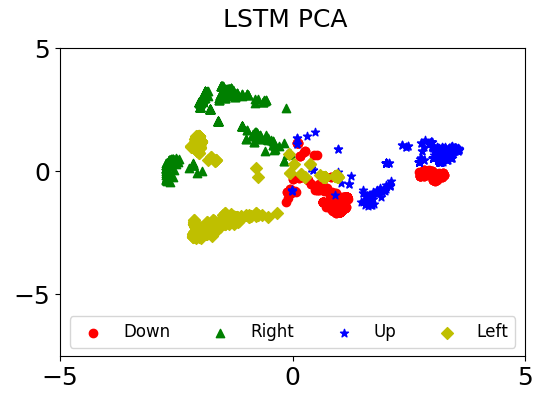}
  \includegraphics[width=122px]{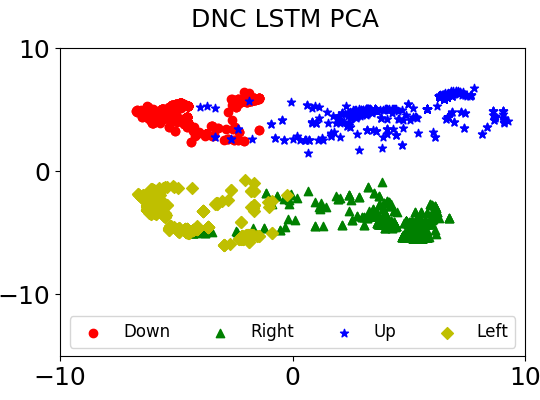}
  \includegraphics[width=122px]{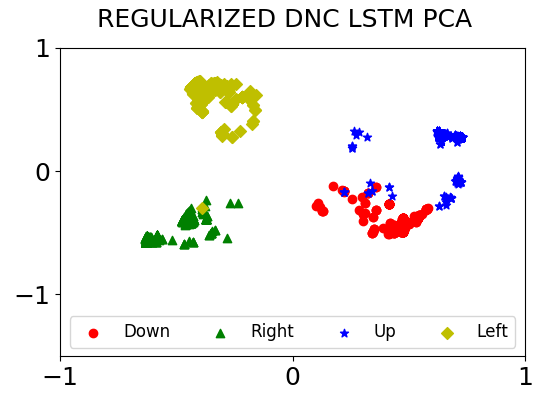}
  \caption{\textbf{\textit{PCA visualization of Last Upstream Layer}}: These visualizations project the $128$ dimensional last upstream space $\Psi(q_i, h_i)$ to $2$ dimensions. The linear readout layer is a perceptron learned in these spaces, and a better neural network separates the different classes with a large margin. Note the varying axis scales for each network, which correspond to the radius of the smallest bounding sphere.}
 \vspace*{-0.15in}
\label{PCA_visualization}
\end{figure*}

Our experiments have two purposes: (1) determine the effect of incorporating neural network memory architectures in robot navigation; and (2) evaluate the predictive capacity of our VC dimension estimates on empirical test error. Our environment is a grid-world involving cul-de-sac obstacles, and we test for each network's ability to interpolate and extrapolate to different length obstacles.

\subsection{Cul-de-sac vs Parallel Walls}
\begin{table}[t!]
	\vspace*{0.05in}
	\begin{center} 
 		\setlength\extrarowheight{10pt}
 		\resizebox{0.48\textwidth}{!}
 		{\begin{tabular}{||c | c | c | c||} 
        \hline
 		\textbf{Parameter} & \textbf{Train} & \textbf{Interpolation Test} & \textbf{Extrapolation Test}\\ [0.6ex] 
 		\hline\hline
 		obs. length (\textit{m}) & 2-20 (even) & 3-19 (odd) & 20-120 (every 1m)\\
 		\hline
 		obs. orientation (\textit{$^{\circ}$}) & 0, 90, 180, 270 & 0, 90, 180, 270 & 0, 90, 180, 270\\
 		\hline
		\end{tabular}}
	\end{center}
\caption{\textbf{\textit{Map generation parameters}}: The interpolation test parameters interpolate obstacle lengths in between the training parameters. The extrapolation test parameters extrapolate to longer obstacle lengths (up to $6\times$ maximum training length).}
\label{table:map_generation_parameters}
\vspace*{-0.25in}
\end{table}

In the grid-world environment, the state $x_t := (\mathsf{x}_t,\mathsf{y}_t)^T$ is the 2D position of the robot. The goal is denoted by $x_{t}^{g} := (\mathsf{x}_t^{g},\mathsf{y}_t^{g})^T$. The robot can take four actions (down, right, up, left), and moves a distance of $1$\;m at each step. The robot cannot turn in place, and starts oriented toward the goal. The robot is equipped with a laser range finder, with a $360 {^\circ}$ field-of-view, that reports relative distance to any perceived obstacles within its field of view. The sensor measurement $z_t$ at time $t$ consists of $N_B = 144$ laser beams with maximum range $5$\;m that report the distance to the closest obstacle along the beam. The obstacle structure is either a cul-de-sac or parallel walls (see Fig.~\ref{fig:U-map}), which the robot cannot determine until it reaches the end of the obstacle.


Neural networks learn to exploit peculiarities in the simulation design. For example, the $A^*$ expert enters in the center of the obstacle and exits near the edges, and we found that the network learns to enter the U-shape if the robot is far away from the edges, and exit if it is near. This behavior is analogous to ``off-loading" memory onto the physical state of the system observed by \cite{DBLP:journals/corr/ZhangLMFA15}, but is not desirable in our case because it does not actually test the network's ability to retain memory. To prevent this memory off-load, we narrow the width of the obstacle to constrain the laser observations to be the same while entering and exiting, thus making the task more challenging by forcing the network to utilize its memory. Table \ref{table:map_generation_parameters} contains the map generation parameters.



\subsection{Neural Networks}
We evaluate four network architectures: FF, LSTM, DNC LSTM, and regularized DNC LSTM. The inputs at each time step are the $144$-dimensional LIDAR reading $z_{t}$, and the $1$-dimensional position heading $atan2(\mathsf{y}_t - \mathsf{y}_t^{g}, \mathsf{x}_t - \mathsf{x}_t^{g})$ representing the heading of the goal from the robot. The FF and LSTM networks have 3 layers of sizes 128 (fully-connected), 128 (fully-connected), and 128 (fully-connected or LSTM respectively). The DNC LSTM network has the same initial structure as the LSTM, in addition to a memory matrix and a fourth fully-connected layer of size 128. The memory matrix is of size $128 \mathsf{x} 32$ and has $2$ read heads and $1$ write head. The original DNC architecture has two final fully-connected layers (LSTM output and memory matrix output) that are summed together. We have converted this architecture into a single linear readout network by combining the LSTM and memory matrix output in the fourth fully-connected layer. This makes the estimation of the VC dimension easier. The regularized DNC LSTM network has the same architecture as the normal DNC LSTM network, except that we include an $L_{2}$ regularization penalty on the parameters that correspond to utilizing or updating the DNC and LSTM. The regularization penalty used is $\lambda=0.1$. The last upstream layer for all $4$ models has dimension $128$.

\subsection{Training Implementation}
The Asynchronous DAgger (Alg.~\ref{alg:algorithm1}) is used to train the neural networks. For the feedforward networks, the training batch size is $5$. For the recurrent networks, the backpropagation through time (BPTT) algorithm is truncated at $5$ steps. We do not backpropagate the external DNC memory parameters through time. We use the RMSProp \cite{tieleman2012lecture} algorithm with a learning rate of $10^{-4}$ to calculate gradients.

\subsection{Results}
We compute three empirical measures of performance: (1) Success rate; (2) classification accuracy; and (3) ratio of path lengths compared to $A^*$. The success rate measures how often the neural network reaches the goal region. The classification accuracy measures of how often the neural network outputs the $A^*$ action. The path length ratio measures the quality of the successful paths versus optimal $A^*$ paths.

In addition, we estimate the VC dimension of our neural networks, using the method described in Section \ref{sec:generalizability}. Our navigation problem is a multi-classification problem. However, we follow the strategy in \cite{Vapnik_1995} and present our results in a one-vs-all binary classification framework. We estimate $\eta_{est}$ by training a linear SVM with a slack penalty ${C=1}$. Table \ref{table:vc-dimension-estimates} presents the training and VC-dimension measures. 

\begin{table}[t!]
	\vspace*{0.05in}
	\begin{center}
    \setlength\extrarowheight{4pt}
	\resizebox{0.48\textwidth}{!}
	{\begin{tabular}{||c|c|c|c|c|c||}
	\hline
	Model (\textit{radius}) & Measure & Down & Right & Up & Left \\   \hline\hline
	\multirow{4}{*}{FF (\textit{37.2})} & $\eta_{est}$ & \textbf{8,558.6} & \textbf{5,740.4}  & \textbf{8,016.7} & \textbf{7,698.9} \\ \cline{2-6} &
    $margin$ & 0.40 & 0.49 & 0.42  & 0.42 \\ \cline{2-6} &
    \textit{nSV} & 340 & 432 & 438 & 335 \\ \cline{2-6} &
    \textit{Training Err. (\%)} & 6.7 & 9.0  & 6.7 & 9.0 \\ \hline 
    \multirow{4}{*}{LSTM (\textit{4.63})} & $\eta_{est}$ & \textbf{165.1} & \textbf{34.8}  & \textbf{48.4} & \textbf{170.1} \\  \cline{2-6}& 
     $margin$ & 0.36 & 0.79 & 0.67  & 0.36 \\ \cline{2-6} &
     \textit{nSV} & 39 & 24 & 30 & 38 \\ \cline{2-6} &
    \textit{Training Err. (\%)} & 0.1 & 0.0 & 0.0  & 0.1 \\ \hline 
    \multirow{4}{*}{DNC LSTM (\textit{9.42})} & $\eta_{est}$ & \textbf{450.3} & \textbf{659.4} & \textbf{1221.4}  & \textbf{386.1} \\  \cline{2-6}& 
     $margin$ & 0.44 & 0.37  & 0.27 & 0.48 \\ \cline{2-6} &
     \textit{nSV} & 33 & 28 & 38 & 26 \\ \cline{2-6} &
    \textit{Training Err. (\%)} & 0.3 & 0.4  & 0.4 & 0.3\\ \hline 
    \multirow{4}{*}{\parbox{2.3cm}{\textbf{\textit{regularized}}\\ \textbf{DNC LSTM (\textit{0.94})}}} & \textbf{$\eta_{est}$} & \textbf{14.1} & \textbf{11.2}  & \textbf{11.3} & \textbf{11.9} \\  \cline{2-6}& 
     $margin$ & 0.25 & 0.28 & 0.28  & 0.27 \\ \cline{2-6} &
     \textit{nSV} & 34 & 17  & 19 & 33 \\ \cline{2-6} &
    \textit{Training Err. (\%)} & 0.5 & 0.0 & 0.0 & 0.3  \\ \hline 
	\end{tabular}}
	\end{center}
\caption{\textbf{\textit{VC dimension estimates}}: \textit{Radius} is the radius of the smallest $L_{2}$ bounding sphere, $\eta_{est}$ is the VC-dimension estimate, \textit{nSV} is the number of support vectors, and \textit{Training Err.} is calculated using the linear SVM. The data set used to estimate these values was generated by sampling 100 episodes using the train map generation parameters. We use $A^*$ to execute the actions in order to generate the same dataset for each architecture, and this particular dataset included 492 down, 588 right, 600 up and 500 left actions. Our method ranks the architectures in the following order: (1) regularized DNC LSTM; (2) LSTM; (3) DNC LSTM; and (4) FF.}
\label{table:vc-dimension-estimates}
\vspace*{-0.25in}
\end{table}

\textbf{\textit{Separability}}: The LSTM, DNC LSTM and regularized DNC LSTM all achieve near-perfect training accuracy, while the FF does not (${\sim}8\%$ error). This result indicates that memory-less neural networks do not have the capacity to correctly separate the action classes and navigate cul-de-sacs. 

\textbf{\textit{Generalization ability}}: Our VC dimension estimates rank the architectures in the following order: (1) regularized DNC LSTM; (2) LSTM; (3) DNC LSTM; and (4) FF. Fig. \ref{PCA_visualization} shows a PCA visualization of the last upstream layer and the neural networks can be viewed as perceptrons separating the classes in these spaces. The prediction performance of $\eta_{est}$ will be evaluated on two data sets: interpolation and extrapolation.

\textbf{\textit{Interpolation}}: The interpolation data set consists of maps with obstacle lengths that are interpolated in between the obstacle lengths encountered in the training set. None of these maps have been encountered in training. Table \ref{table:interpolation} displays the interpolation results and shows that all 3 memory networks are able to successfully generalize. For the FF network, it is interesting to note that the $A^*$ ratio is greater than 1 for parallel walls, and less than 1 for cul-de-sacs. This pattern indicates that the FF network turns around before it reaches the end of the obstacle length, which is not desirable.

\textbf{\textit{Extrapolation}}: The extrapolation data set consists of maps with obstacle lengths that range from $20$ to $120$\;m, which corresponds to $6\times$ the length of the maximum obstacle length encountered during training. Table \ref{table:extrapolation} ranks the architectures in the following order: (1) regularized DNC LSTM; (2) LSTM; (3) DNC LSTM; and (3) FF, confirming the predictions made by our VC dimension estimates. The regularized DNC LSTM is able to generalize almost perfectly. The LSTM is able to successfully complete all episodes, but it exhibits the same pattern in the $A^*$ ratio that indicates it is turning around before it reaches the end of the obstacle. The DNC LSTM exhibits similar behavior, but is only able to successfully complete ${\sim}55\%$ of the episodes. These results indicate that the LSTM and DNC LSTM have overfit the training set, with the LSTM generalizing slightly better. 

This relative degree of overfitting is reasonable, since the DNC LSTM has strictly more ``memory parameters" than the LSTM. Overfitting is expected to occur as our networks are complex, yet the amount of memory to actually navigate cul-de-sac environments is relatively low. Indeed, we only need 3 states that determine when the end of the obstacle has been reached, whether the end is closed or open, and when the robot has exited the entrance of the obstacle. It is thus not surprising that regularization greatly improves the quality of the learned network.

\begin{figure}[t!]
\vspace*{0.05in}
\centering
\begin{subfigure}{0.5\textwidth}
\centering
\includegraphics[width=114px]{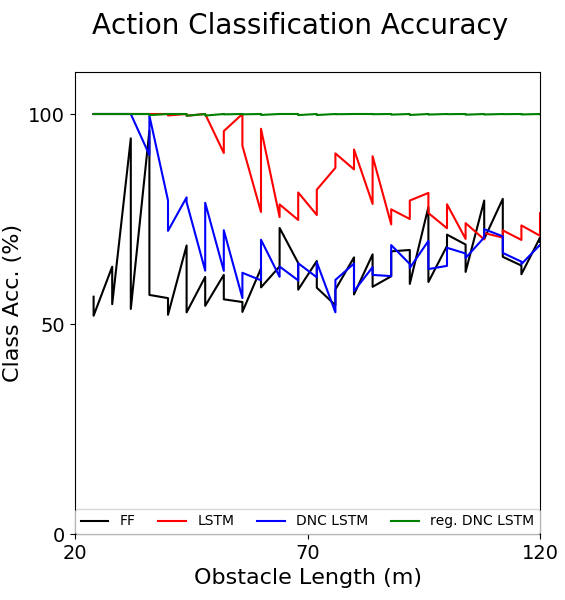}
\includegraphics[width=110px]{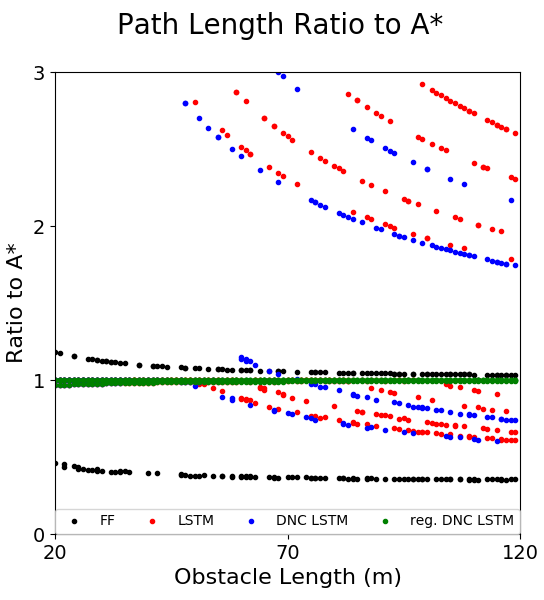}
\vspace{0.0cm}
\end{subfigure}
\caption{\textit{\textbf{Left}: Classification Accuracy performance as obstacle length increases}. \textit{\textbf{Right}: Ratio to $A^*$ as obstacle length increases}. }
\label{fig:extrapolation_performance_graphs}
\vspace*{-0.2in}
\end{figure}

Figure \ref{fig:extrapolation_performance_graphs} shows the classification accuracy and $A^*$ path length ratio of all networks against obstacle length. The regularized DNC LSTM has $100\%$ classification accuracy and an $A^*$ ratio of 1 for all obstacle lengths, indicating \emph{perfect generalization}. The DNC LSTM has perfect classification accuracy up to obstacle lengths of $30$\;m ($1.5\times$ maximum training length), while the LSTM has perfect classification up to obstacle lengths of $50$\;m ($2.5\times$ maximum training length). 

\begin{table}[t!]
	\vspace*{0.0501in}
	\begin{center}
    \setlength\extrarowheight{4pt}
	\resizebox{0.48\textwidth}{!}
	{\begin{tabular}{||c|c|c|c|c||}
	\hline
	Model & Map Type & \textit{Success ($\%$)} & \textit{Class. Acc ($\%$)} & \textit{$A^{*}$ Ratio} \\   \hline\hline
	\multirow{2}{*}{FF} & \textit{Parallel Walls} & 84.0 & 61.9 & 1.194 \\ \cline{2-5} &
    \textit{Cul-de-sac} & 52.0 & 54.1 & 0.709 \\ \hline
    \multirow{2}{*}{LSTM} & \textit{Parallel Walls} & 100.0 & 100.0 & 1.000 \\  \cline{2-5}& 
     \textit{Cul-de-sac} & 100.0 & 98.9 & 1.011 \\ \hline
    \multirow{2}{*}{DNC LSTM} & \textit{Parallel Walls} & 100.0 & 100.0 & 1.000 \\  \cline{2-5} & 
     \textit{Cul-de-sac} & 100.0 & 98.6 & 1.020 \\ \hline
    \multirow{2}{*}{\parbox{1.6cm}{\textit{regularized}\\ DNC LSTM}} & \textit{Parallel Walls} & 100.0 & 100.0 & 1.000 \\  \cline{2-5}& 
     \textit{Cul-de-sac} & 100.0 & 98.9 & 1.013 \\ \hline
	\end{tabular}}
	\end{center}
\caption{\textbf{\textit{Interpolation Generalization}}: This data set consists of 100 maps sampled from interpolation map generation parameters and is shared across all of the networks. Episodes over $200$ time steps were terminated and counted as a failure.}
\label{table:interpolation}
\vspace*{-0.02in}
\end{table}

\begin{table}[t!]
	\vspace*{-0.2cm}
	\begin{center}
    \setlength\extrarowheight{4pt}
	\resizebox{0.48\textwidth}{!}
	{\begin{tabular}{||c|c|c|c|c||}
	\hline
	Model & Map Type & \textit{Success ($\%$)} & \textit{Class. Acc ($\%$)} & \textit{$A^{*}$ Ratio} \\   \hline\hline
	\multirow{2}{*}{FF} & \textit{Parallel Walls} & 75.5 & 65.1 & 1.041 \\ \cline{2-5} &
    \textit{Cul-de-sac} & 47.5 & 58.9 & 0.375 \\ \hline
    \multirow{2}{*}{LSTM} & \textit{Parallel Walls} & 100.0 & 76.8 & 1.660 \\  \cline{2-5}& 
     \textit{Cul-de-sac} & 100.0 & 84.1 & 0.891 \\ \hline
    \multirow{2}{*}{DNC LSTM} & \textit{Parallel Walls} & 56.5 & 65.9 & 1.781 \\  \cline{2-5} & 
     \textit{Cul-de-sac} & 54.5 & 69.7 & 0.892 \\ \hline
    \multirow{2}{*}{\parbox{1.6cm}{\textbf{\textit{regularized}}\\ \textbf{DNC LSTM}}} & \textbf{\textit{Parallel Walls}} & \textbf{100.0} & \textbf{100.0} & \textbf{1.000} \\  \cline{2-5}& 
     \textbf{\textit{Cul-de-sac}} & \textbf{100.0} & \textbf{99.9} & \textbf{0.994} \\ \hline
	\end{tabular}}
	\end{center}
\caption{\textbf{\textit{Extrapolation Generalization}}: This data set consists of 2 maps at each obstacle length between $20-120$\;m, resulting in 200 total maps. The remaining map parameters were sampled from the extrapolation map generation parameter. Episodes over $500$ time steps were terminated and counted as a failure. }
\label{table:extrapolation}
\vspace*{-0.2in}
\end{table}

For obstacle lengths $<50$\;m, the $A^*$ path length ratio graph shows that the DNC LSTM and LSTM are matching the $A^*$ paths. With longer obstacles, we see 2 regimes of points. The top regime corresponds to the path lengths in the parallel wall scenarios, while the bottom regime corresponds to the cul-de-sac scenario. In addition, notice the downward pattern in these points. 

If the robot almost reaches the end of the obstacle before turning around, it will have entered the obstacle, exited the obstacle, and gone down the side of the obstacle ({$1+1+1=3$}). Thus, the $A^*$ path ratio should be ${\sim}3\times$ and ${\sim}1\times$ for the parallel wall and cul-de-sac environments, respectively. If the robot turns around halfway, it would have entered halfway into the obstacle, exited halfway, and gone down the side of the obstacle ({$\frac{1}{2} + \frac{1}{2} + 1 = 2$}). As a result, the $A^*$ path ratio should be ${\sim}2\times$ for the parallel wall and ${\sim}\frac{2}{3}\times$ for the cul-de-sac. The downward trend in the ratio as the length increases indicate that in successful episodes, the DNC LSTM and LSTM seems to always turn around at $60$-$80$\;m. 
Likewise, we also see 2 regimes of points for the FF $A^{*}$ path ratios centered around $1\times$ and $1/3\times$. This pattern reveals that when the FF network is successful, it entirely bypasses the obstacle. This is also not desirable because the network is just exploiting another peculiarity in the simulation design. The attached video visualizes these behaviors.

\subsection{Discussion} Our empirical test results match the predictions made by our VC dimension estimation method, indicating that our method is a good indicator of generalization. One benefit of our method is that it has a clear measure of generalization $\eta_{est}$. While a simple classification problem may have a clear measure of performance, in our sequential prediction problem, we presented 3 measures of performance which all captured different behaviors of the network. It is unclear which one is the most desirable and should be optimized.

More importantly, only the extrapolation data set was able to differentiate the models in our experiments. All the memory models had the same performance on the interpolation data set, highlighting the second benefit of our method.
\begin{remark*}
Assessing generalization ability from empirical test sets is dependent on the choice of training set and the testing set, while our VC dimension estimation method is dependent only on the choice of the training set.
\end{remark*}

The standard evaluation method in deep learning uses a held-out test set that is randomly selected from the training set. This choice of test set is similar to our interpolation test set choice. The test error on a poorly chosen test set may not provide a good metric of generalization because the distribution is too similar to that of the train set. These observations highlight the benefits of complementing current empirical test measures with our VC dimension estimates.

\section{Conclusion}
\label{sec:conclusion}
This paper considered the problem of learning closed-loop perception-action policies for autonomous robot navigation. Unlike traditional feedback motion planning approaches that rely on accurate global maps, our approach can infer appropriate actions directly from sensed information by using a neural network policy representation. We argued that including memory in the network structure is fundamental for summarizing past information and achieving good performance as measured by its ability to separate the correct action from other choices and to generalize to unseen environments. Our main contribution is a method for estimating the VC dimension of the last network layer (after all upstream layer transformations) that can be used as an accurate generalization ability measure that depends only on the training set choice. Finally, we proposed a new parallel training algorithm for supervised learning of closed-loop policies in sequential prediction problems. Our analysis and results demonstrated the need for and superiority of including external memory when increasing the depth of the cul-de-sacs present in the environment.

Future work will focus on transfering learned perception-action policies to a real robot and evaluating the regret against simulataneous mapping and planning algorithms in a physical environment. We are also interested in extending the approach for robot systems with higher state, control, and measurement dimensions such as velocity-actuated ground robots in $SE(2)$ and force-actuated aerial robots in $SE(3)$ using 3-D depth sensors or image-based information.



\balance

\bibliographystyle{ref/IEEEtran}
\bibliography{ref/ref.bib}
\end{document}